\documentclass[conference, compsocconf]{IEEEtran}
\usepackage[cmex10]{amsmath}
\usepackage{url}
\usepackage{tikz-qtree}
\usepackage{graphicx}
\usepackage{caption}
\usepackage{fixltx2e}
\usepackage{float}
\usepackage{mathtools}
\usepackage[utf8]{vietnam}
\usepackage[english]{babel}

\DeclareMathOperator*{\argmax}{arg\,max}

\DeclareMathOperator*{\softmax}{softmax}
\DeclareMathOperator*{\length}{length}

\begin{document}

\title{On the Use of Machine Translation-Based Approaches for Vietnamese Diacritic Restoration}

\author{\IEEEauthorblockN{Thai-Hoang Pham}
\IEEEauthorblockA{Alt Inc\\
Hanoi, Vietnam\\
Email: \textit{phamthaihoang.hn@gmail.com}}
\and
\IEEEauthorblockN{Xuan-Khoai Pham}
\IEEEauthorblockA{FPT University\\
Hanoi, Vietnam\\
Email: \textit{khoaipxmse0060@fpt.edu.vn}}
\and
\IEEEauthorblockN{Phuong Le-Hong}
\IEEEauthorblockA{Vietnam National University\\
Hanoi, Vietnam\\
Email: \textit{phuonglh@vnu.edu.vn}}
}

\maketitle

\begin{abstract}
This paper presents an empirical study of two machine translation-based
approaches for Vietnamese diacritic restoration problem, including
phrase-based and neural-based machine translation models. This is the
first work that applies neural-based machine
translation method to this problem and gives a thorough comparison to
the phrase-based machine translation method which is
the current state-of-the-art method for this problem. On a large
dataset, the phrase-based approach has an accuracy of $97.32\%$ while
that of the neural-based approach is $96.15\%$. While the neural-based
method has a slightly lower accuracy, it is about twice faster than the
phrase-based method in terms of inference speed. Moreover, neural-based
machine translation method has much room for future improvement such as
incorporating pre-trained word embeddings and collecting more training
data.
\end{abstract}

\IEEEpeerreviewmaketitle

\section{Introduction}
Vietnamese and many other languages that use Roman characters have
diacritic marks. Normally, we compound diacritic marks with syllables
to form a meaningful word in writing.  Recently, with the availability
of the electronic texts such as email and message, people
often remove diacritic from words, eitheir due to their intention of
speed typing or their unfamiliarity with Vietnamese input methods. In particular,
people need to install some Vietnamese keyboard applications and
follow some typing rules such as Telex, VNI, or VIQR to type
Vietnamese texts with diacritic marks. According to the statistics
in~\cite{Do:2013}, $95\%$ Vietnamese words contain diacritic marks and
$80\%$ of these words are ambiguous when removing diacritic marks from
them. Thus, reading non-diacritic is difficult for both human and
machine. For example, a non-diacritic sentence ``\textit{Co ay rat dam dang}''
can be interpreted as ``\textit{Cô ấy rất đảm đang}'' (She is very capable) or
``\textit{Cô ấy rất dâm đãng}'' (She is very lustful).  

Vietnamese diacritic marks appear at all vowel characters and one
consonant character. There are two types of diacritic marks for
Vietnamese. One type (Type-1) is added to a character to transform
this character to another one, and another type (Type-2) is used to
change the tone of a word. Table~\ref{tab:0} shows a map from
non-diacritic characters to diacritic characters.

Several approaches have been proposed to restore diacritics marks for
Vietnamese such as rule-based, dictionary-based, and machine
learning-based methods. Recently, machine translation-based method has
emerged as the best solution for this problem. The idea of this method
is treating non-diacritic texts and diacritic texts as source and
target languages in machine translation formulation. Removing
diacritic marks from regular texts is a trivial task, so it is easy to
create a bi-lingual dataset for this problem. Finally, several machine
translation toolkits are trained on this dataset to learn how to
translate from non-diacritic texts to diacritic texts.

In this paper, we present a thorough empirical study of applying
common machine translation approaches for Vietnamese diacritic
restoration problem including phrase-based and neural-based machine
translation methods. Our work is the first study that not only investigates
the impact of neural-based machine translation to Vietnamese diacritic
restoration problem but also compares the strengths and weaknesses of
these two approaches for this problem. We also conduct experiments on
a large dataset that consists of about 180,000 sentence pairs to get
reliable results. In summary, the accuracy scores of the phrase-based
and neural based methods are $97.32\%$ and $96.15\%$ on our dataset
respectively which are state-of-the-art results for this problem.

\begin{table}[t]
\center
\caption{A map from non-diacritic characters to diacritic characters}
\resizebox{\linewidth}{!}{
\begin{tabular}{|l|l|l|}
\hline 
Non-diacritic & Non-diacritic + Type 1 & Non-diacritic + Type 1 + Type 2 \\ 
\hline 
a & a, ă, â & a, à, ả, ã, ạ, ă, ắ, ằ, ẳ, ẵ, ặ, â, ấ, ầ, ẩ, ẫ, ậ \\ 
\hline 
e & e, ê & e, é, è, ẻ, ẽ, ẹ, ê, ế, ề, ể, ễ, ệ \\ 
\hline 
i & i & i, í, ì, ỉ, ĩ, ị \\ 
\hline 
o & o, ô & o, ó, ò, ỏ, õ, ọ, ô, ố, ồ, ổ, ỗ, ộ \\ 
\hline 
u & u, ư & u, ú, ù, ủ, ũ, ụ, ư, ứ, ừ, ử, ữ, ự \\ 
\hline 
y & y & y, ý, ỳ, ỷ, ỹ, ỵ \\ 
\hline 
d & d, đ &  \\ 
\hline 
\end{tabular}}
\label{tab:0} 
\end{table}

The remainder of this paper is structured as
follows. Section~\ref{sec:relatedwork} summarizes related work on
Vietnamese diacritic restoration. Section~\ref{sec:method} describes
two common machine translation approaches for this task including
phrase-based and neural-based methods. Section~\ref{sec:experiment}
gives experimental results and discussions. Finally,
Section~\ref{sec:conclusion} concludes the paper.

\section{Related Work}\label{sec:relatedwork}
There are several methods to automatically restore diacritics
from non-diacritic texts which are divided into two main approaches. The first approach is characterized by the
use of dictionaries and rule sets. The performance of this approach is
heavily dependent on qualities of pre-compiled dictionaries and rules, and
domains of texts. In particular,
VietPad\footnote{\url{http://vietpad.sourceforge.net/}}, a Vietnamese
Unicode text editor, uses a dictionary that stores most Vietnamese
words to 1-to-1 map from non-diacritic words to diacritics words. This
method is not effective because many words in the dictionary do 
not appear frequently in real texts. The accuracy of this tool is
about from $60\%$ to $85\%$ depending on the domain of
texts. VietEditor\footnote{\url{http://lrc.quangbinhuni.edu.vn:8181/dspace/bitstream/TVDHQB_123456789/264/3/Themdautiengviet.pdf}
  (Vietnamese)} toolkit alleviate the weakness of VietPad by building the
phrase dictionary and uses it after mapping words to find the most
appropriate outputs.

The second approach is using machine learning methods to handle this problem. They apply some common machine learning models
such as conditional random field (CRF), support vector machine (SVM),
and N-gram language models to restore diacritics for Vietnamese
texts.~\cite{Truyen:2008} proposes viAccent toolkit that is a
combination of N-grams, structured perception and CRF. N-grams is used
as features for CRF to label diacritics for input sentences. They
achieve an accuracy of $94.3\%$ on a newspaper
dataset.~\cite{Trong:2009} combines both language model and
co-occurrence graph to capture information from non-diacritic
texts. For inference, they apply dynamic programming to find
the best output sequences based on information from input texts.~\cite{Luu:2012} proposes a pointwise approach for automatically
recovering diacritics, using three features for classification
including N-grams of syllables,  syllable types, and dictionary word
features. They achieve an accuracy of $94.7\%$  by using SVM
classifier.~\cite{Nguyen:2010} gives an empirical study for Vietnamese
diacritic restoration by investigating five strategies: learning from
letters, learning from semi-syllables, learning from syllables,
learning from words, and learning from bi-grams. They combine AdaBoost
and C4.5 algorithms to get better results. Their best accuracy is
$94.7\%$ when using letter-based feature set.~\cite{Nguyen:2012}
formulates this task as sequence tagging problem and use CRF and SVM
models to restore diacritics. They achieve the accuracy of $93.8\%$ on
written texts by using CRF at syllable level.

Machine translation-based approach has emerged as the best way
to handle this problem.~\cite{Do:2013} and~\cite{Pham:2013} formulate
this task as machine translation problem and apply phrase-based
translation method by using Moses toolkit. Both of them report the
accuracy of $99\%$ on their dataset but the size of this dataset is
relativity small. For this reason, in this paper, we experiment this
method on a large dataset to get fair comparisons.

\section{Methodology}\label{sec:method}
We treat the diacritic restoration problem as a machine translation problem and
apply phrase-based and neural-based machine translation methods for this task. In particular, non-diacritic and
diacritic texts are considered as source and target languages
respectively, and machine translation models are trained to learn how
to restore diacritics.

\subsection{Phrase-Based Machine Translation}
Phrase-based machine translation is one type of statistical machine
translation that translate phrases in source language to phrases in
target language~\cite{Koehn:2003}, \cite{Koehn:2007}. The main idea of
this approach is an input sentence is segmented into a number of
sequences of consecutive words (phrases). After that, each phrase in
source language is translated to one phrase in target language that
might be reordered. 

The phrase translation model is based on the noisy channel model. In
particular, it tries to maximize the translation probability from
source sentence $\textbf{f}$ to target sentence $\textbf{e}$. Applying
Bayes rule, we can reformulate this probability as  
\begin{equation}
\underset{\textbf{e}}{\argmax} \; p(\textbf{e}|\textbf{f}) =
\underset{\textbf{e}}{\argmax} \; p(\textbf{f}|\textbf{e})
p(\textbf{e}) \omega^{\length(\textbf{e})}
\end{equation}
where $\omega$ is added to calibrate the output length.

During decoding, the source sentence $\textbf{f}$ is segmented into a
sequence of $N$ phrases $\textbf{f}_{i}$. After that, each source
phrase $\textbf{f}_{i}$ is translated to target phrase
$\textbf{e}_{i}$ by the probability distribution
$\phi(\textbf{f}_{i}|\textbf{e}_{i})$. The sequence of target phrases
might be reorderd by a relative distortion probability distribution
$d(start_{i}, end_{i-1})$ where $start_{i}$ denotes the start position
of the source phrase that was translated into the $i^{th}$ target
phrase, and $end_{i-1}$ denotes the end position of the source phrase 
that was translated into the $(i-1)^{th}$ target phrase. As sum, $p(\textbf{f}|\textbf{e})$ can be calculated as
\begin{equation}
\prod_{i=1}^{N}\phi(\textbf{f}_{i}|\textbf{e}_{i})d(start_{i},
end_{i-1}) 
\end{equation}

\begin{figure}[h]
\centering
\resizebox{0.8\linewidth}{!}{
\includegraphics{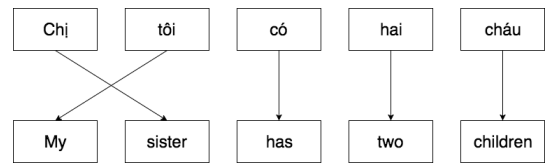}
}
\caption{Phrase-based machine translation system that translates
  source sentence ``\textit{Chị tôi có hai cháu}'' to target sentence ``\textit{My sister
  has two children}''} 
\label{fig:1}
\end{figure}

Figure~\ref{fig:1} presents an example of phrase-based machine translation system.

\subsection{Neural-Based Machine Transslation}
In the last few years, deep neural network approaches have achieved
state-of-the-art results in many natural language processing (NLP)
task. There are a lot of research that applied deep learning methods
to improve performances of their NLP systems. For machine translation
problem,~\cite{Cho:2014},~\cite{Luong:2015} proposed a
sequence-to-sequence model that achieved best results for many
bi-lingual translation tasks. The general architecture of this model
is the combination of two recurrent neural networks. One network encodes
a sequence of words in source language into a fixed-length vector
representation, and the other decodes this vector into another
sequence of words in target language. Both of these two networks are
jointly trained to maximize the conditional probability of a target
sentence given a source sentence. In particular, the conditional
probability $p(\textbf{e}|\textbf{f})$ is computed as
\begin{equation}
\log (\textbf{e}|\textbf{f}) = \sum_{j=1}^{m} \log p(\textbf{e}_{j}|\textbf{e}_{<j}, \textbf{s})
\end{equation}
where $\textbf{s}$ is representation vector produced by encoder module.

\begin{figure}[h]
\centering
\resizebox{0.8\linewidth}{!}{
\includegraphics{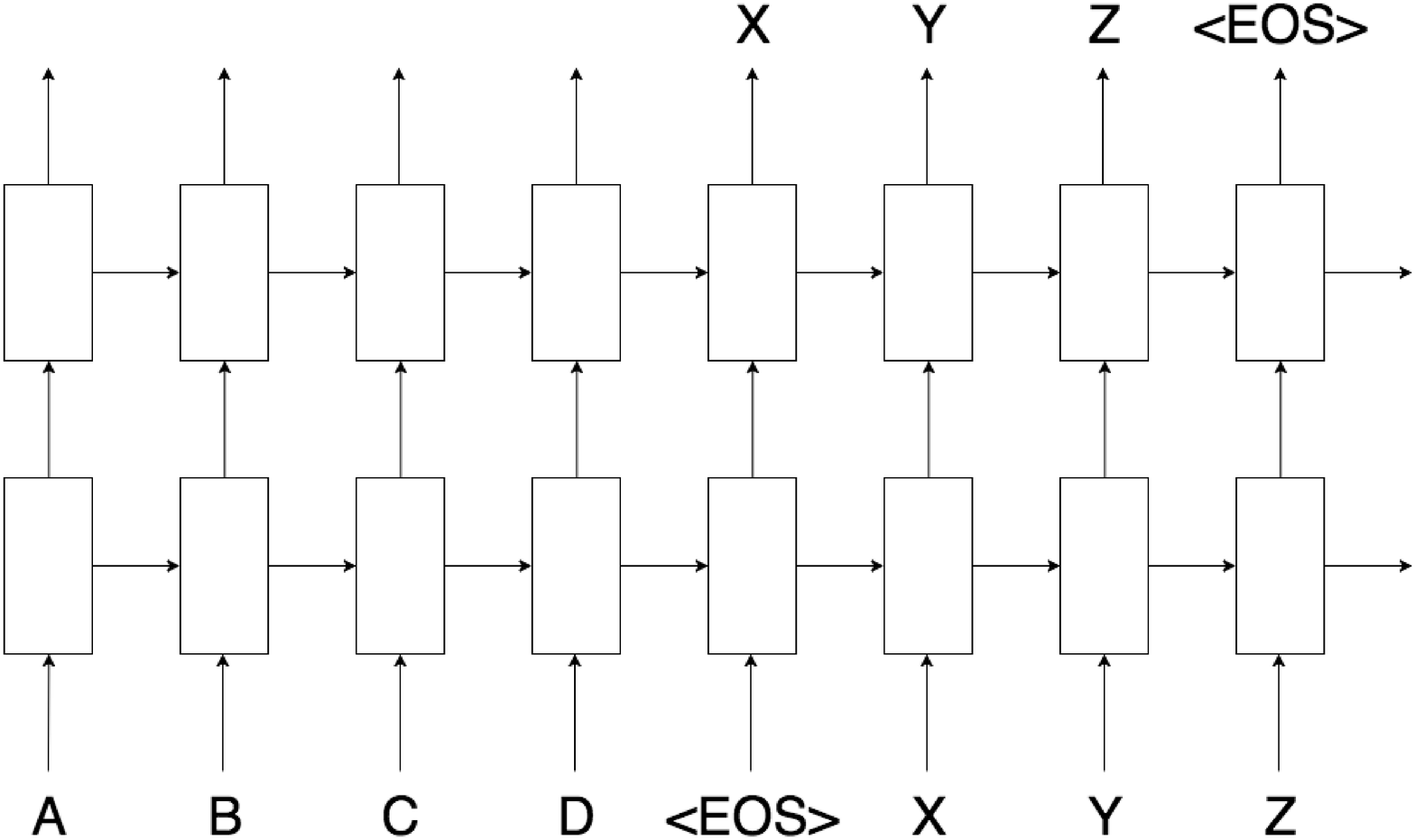}
}
\caption{The architecture of sequence-to-sequence model}
\label{fig:2}
\end{figure}

In decoding stage, the conditional probability of a word given previous words is computed as
\begin{equation}
p(\textbf{e}_{j}|\textbf{e}_{<j}, \textbf{s}) = \softmax(g(\textbf{h}_{j}))
\end{equation}
where $\textbf{h}_{j}$ is the hidden state at time step $j$ of
recurrent neural network that computed by previous hidden state and
representation vector $\textbf{s}$, and $g$ is a function that
transforms the hidden state to vocabulary-sized
vector. Figure~\ref{fig:2} present the architecture of
sequence-to-sequence model. 

\section{Experiment}\label{sec:experiment}

\subsection{Dataset}
To evaluate these machine translation approaches for this problem on
the large dataset, we first collect 10,000 news articles from the web
and then remove non-standard characters and diacritics from the
original text to build a parallel corpus of about 180,000 sentence pairs. We use $80\%$ of this
dataset as a traning set, $10\%$ of this dataset as a development set,
and the remaining as a testing set. Table~\ref{tab:1} shows the
statistics of this dataset.

\begin{table}[h]
\center
\caption{Numbers of sentences and words in each part of the dataset}
\begin{tabular}{|l|l|l|}
\hline 
 & \#sentence & \#word \\ 
\hline 
Training & 140474 & 3391193 \\ 
\hline 
Development & 17559 & 424309 \\ 
\hline 
Testing & 17559 & 423729 \\ 
\hline 
\end{tabular} 
\label{tab:1} 
\end{table}
\subsection{Evaluation Method}
We utilize Moses\footnote{\url{http://www.statmt.org/moses/}} and
OpenNMT\footnote{\url{http://opennmt.net/}} toolkits as
representatives for phrase-based and neural-based machine translation
approaches respectively. To evaluate performaces of these toolkits, we
use an accuracy score that calculates the percentage of correct words
restored by these systems and a BLEU score that evaluates results of
translation systems.

\subsection{Results and Discussions}
We train these two toolkits on the training set and use the
development set for tuning. In particular, we use the development set
to adjust parameters of Moses toolkit and to early stop the training
of OpenNMT toolkit.
Finally, we restore diacritics of texts
in the testing set. We use the standard setting in these two toolkits
when traning and inference. For Moses toolkit, we use
KenLM\footnote{\url{http://kheafield.com/code/kenlm/}} to build 3-gram
language model and
GIZA++\footnote{\url{http://www.statmt.org/moses/giza/GIZA++.html}}
for word alignment. For OpenNMT toolkit, we use the
sequence-to-sequence model described in~\cite{Luong:2015}. This model consists of encoder and decoder modules that are
recurrent neural network models. Table~\ref{tab:2} shows the accuracy
and BLEU scores of these two systems.  
\begin{table}[h]
\center
\caption{Accuracy and BLEU scores of each system on our testing set}
\begin{tabular}{|l|l|l|}
\hline 
 & Accuracy & Bleu \\ 
\hline 
Phrase-based (Moses) & 97.32 & 94.11 \\ 
\hline 
Neural-based (OpenNMT) & 96.15 & 91.59 \\ 
\hline 
\end{tabular} 
\label{tab:2} 
\end{table}

The main purpose of BLEU score is evaluating the quality of the
machine translation system~\cite{Papineni:2002}. It is not suitable to
use this score to assess the performances of these systems for
diacritic restoration task. We, therefore, focus only the accuracy
score. Both of these systems achieve the state-of-the-art results for
Vietnamese diacritic restoration task. In particular, the Moses
toolkit achieves an accuracy of $97.32\%$, which is slightly higher
than an accuracy of $96.15\%$ of OpenNMT toolkit. The reason for this
result may be the size of the training set. Neural-based approach
often requires a large amount of training data to get a good
performance while our training set has only 140,000 sentence
pairs. Moreover, previous works show that using pre-trained word
embeddings help to improve greatly the performance of the neural
machine translation, but in this task, we do not use any pre-trained
word embeddings.  

While the performance of OpenNMT toolkit is not better than Moses
toolkit in this dataset, we realize that OpenNMT toolkit requires less
time for training and has a higher speed when restore diacritics for
input sentences. Table~\ref{tab:3} shows the training time and the
average speed at the inference stage of these two systems. 
\begin{table}[h]
\center
\caption{Training times (hours) and Testing speeds (\#sentence/second) of Moses and OpenNMT toolkits}
\begin{tabular}{|l|l|l|}
\hline 
 & Training & Testing \\ 
\hline 
Phrase-based (Moses) & 12 hours & 10 sent/s \\ 
\hline 
Neural-based (OpenNMT) & 8 hours & 22 sent/s \\ 
\hline 
\end{tabular} 
\label{tab:3} 
\end{table}

In particular, we train and evaluate two system at the same setting. The details of hardware configuration are Intel Xeon E5-2686, 60GB of RAM, and Tesla K80 12GB. OpenNMT toolkit takes 8 hours for training while Moses toolkit needs 12 hours. For inference stage, OpenNMT toolkit is likely to handle 22 input sentences per second which is twice as fast as Moses toolkit. The reason is that OpenNMT toolkit can take advantage of the performance of GPU that has many CUDA cores to parallel processing.
\section{Conclusion}\label{sec:conclusion}
In this paper, we present the empirical study of machine
translation-based approaches for Vietnamese diacritic restoration. In
particular, we conduct experiments to compare two common approaches
for machine translation including phrase-based and
neural-based methods. Both of two systems achieve
state-of-the-art results for Vietnamese diacritic restoration
task. While the phrase-based method has a slightly higher accuracy,
the neural-based method requires less time for training and has much
faster inference speed.

In the future, on the one hand, we plan to improve
neural-based approach by enlarging our corpus so as to provide more
data for training. On the other hand, we would like to incorporate 
pre-trained Vietnamese word embeddings to boost the accuracy of this
approach.
\bibliographystyle{IEEEtran}
\bibliography{IEEEabrv,mybib}

\end{document}